\documentclass[runningheads]{llncs}

\usepackage[T1]{fontenc}
\usepackage[utf8]{inputenc}
\usepackage{graphicx}
\usepackage{subcaption}   
\usepackage{amsmath,amssymb,amsfonts}

\usepackage{amsthm}
\usepackage{booktabs}
\usepackage{multirow}   
\usepackage{times}
\usepackage{soul}
\usepackage{url}
\usepackage[hidelinks]{hyperref}
\usepackage[switch]{lineno}
\usepackage[ruled]{algorithm2e}
\usepackage{diagbox}
\usepackage{arydshln}
\usepackage{bm}
\usepackage{verbatim}
\usepackage[numbers]{natbib}
\usepackage{float}

\begin{document}

\title{MRAFnd: Multimodal Retrieval-Augmented Framework for Zero-Shot Fake News Detection}

\titlerunning{MRAFnd: Multimodal Retrieval-Augmented Framework}

\author{
\begin{minipage}[t]{\textwidth}
\centering
Lehan Zhang\inst{1} \and
Yinlei Cheng\inst{1} \and
Shiqi Hu\inst{2} \\[2pt]
Yiheng Zhou\inst{1} \and
Shangxi Li\inst{1} \and
Naidong Zhao\inst{1}\thanks{This study was supported by College Students' Innovative Entrepreneurial Training Plan Program}
\end{minipage}
}
\authorrunning{Lehan Zhang et al.}
\institute{
\begin{tabular}{c}
\inst{1} Beijing Institute of Fashion Technology, Beijing, China \\
\inst{2} Shenyang University of Technology, Shenyang, China \\
\email{sxyznd@bift.edu.cn}
\end{tabular}
}

\maketitle

\begin{abstract}
The rapid dissemination of multimodal content has intensified the spread of fabricated news, presenting a substantial threat to social integrity. A formidable challenge for current detection systems is identifying misinformation related to novel events in zero-shot scenarios. Prevailing zero-shot methods typically assess news items in isolation via semantic matching, a strategy that fails to recognize the recycled disinformation tactics from past campaigns and lacks the sophisticated reasoning needed to identify subtle, cross-modal discrepancies. To surmount these deficiencies, we introduce \textbf{MRAFnd}, a novel \underline{\textbf{M}}ultimodal \underline{\textbf{R}}etrieval-\underline{\textbf{A}}ugmented Framework for Zero-Shot \underline{\textbf{F}}ake \underline{\textbf{N}}ews \underline{\textbf{D}}etection. MRAFnd emulates a collaborative team of analysts to verify news veracity. The framework initiates with \textbf{Multimodal Similarity-based News Retrieval} to assemble a corpus of contextually analogous articles from an unlabeled reference database. Subsequently, during the \textbf{Bifurcated Evidential Reasoning} stage, agents perform a dual-directional analysis to extract critical patterns from the retrieved evidence. Finally, a \textbf{Multi-Agent Collaborative Debate}, involving Analyst and Arbiter agents, engages in a structured discourse to arrive at a definitive and robust conclusion. Comprehensive experiments on three benchmark datasets reveal that MRAFnd markedly surpasses state-of-the-art baselines, achieving an accuracy gain of up to 2.35\% on the demanding Weibo-21 dataset.

\end{abstract}

\section{Introduction}

The ascent of social media has revolutionized information dissemination, establishing multimodal content—integrating text and imagery—as the principal medium of communication. This paradigm shift, however, has concurrently accelerated the propagation of multimodal fake news, which deceptively combines fabricated text with authentic or manipulated visuals to mislead audiences \cite{zhou2020survey, tong2024mmdfnd}. Such disinformation can undermine public trust, sway public opinion, and provoke social instability, thus posing a grave risk to societal cohesion. Consequently, developing automated and precise methods for detecting multimodal fake news has emerged as a paramount research imperative. Existing detection models frequently fall short in adapting to the fluid landscape of fake news, where new narratives and manipulated content concerning emergent events surface daily, necessitating constant model adaptation in data-deficient, zero-shot contexts \cite{giachanou2022survey}.

Current research in multimodal fake news detection is primarily divided into two streams. The first stream includes supervised deep learning models, which have demonstrated significant success when trained on extensive annotated datasets \cite{wang2018eann, khattar2019mvae, lu2025dammfnd, tong2025dapt}. These models generally utilize advanced fusion techniques to identify cross-modal inconsistencies. Their efficacy, however, is fundamentally contingent on the availability of large-scale labeled data, making them unsuitable for confronting fake news about novel events where no prior annotations exist. The second stream aims to overcome this data dependency using few-shot or zero-shot learning approaches. While promising, these methods often depend on superficial semantic matching and are deficient in the sophisticated reasoning capabilities essential for uncovering the nuanced contextual and logical incongruities that define advanced disinformation \cite{qian2021hierarchical}.

A core limitation of contemporary zero-shot methodologies is their tendency to analyze a target news article as an isolated entity. This approach neglects a critical characteristic of disinformation campaigns: despite addressing new events, fake news articles frequently repurpose narrative frameworks, image manipulation techniques, or stylistic formats from earlier campaigns. For example, a fabricated news story might pair an image of a political rally with a misleading caption. Although the specific event is new, the underlying tactic of misrepresenting crowd size or event context is a recurrent trope. Identifying such deceit requires more than a surface-level analysis of the image and text; it demands a profound, comparative reasoning process that situates the article within a broader context of similar real and fake news instances. \textbf{An isolated examination is thus insufficient for exposing these recycled, malicious patterns.}

To address these challenges, we introduce a new zero-shot framework: \textbf{MRAFnd}, a \underline{\textbf{M}}ultimodal \underline{\textbf{R}}etrieval-\underline{\textbf{A}}ugmented Framework for Zero-Shot \underline{\textbf{F}}ake \underline{\textbf{N}}ews \underline{\textbf{D}}etection. Inspired by collaborative investigative journalism and the advanced reasoning of Multimodal Large Language Models (MLLMs), MRAFnd simulates a team of analysts collaborating to verify a suspicious news article. The framework's operation involves: 1) Retrieving a set of similar, contextually relevant news articles from a vast, unannotated corpus to establish an evidentiary baseline. 2) Deploying a pair of LLM agents to conduct a bifurcated analysis of this evidence, extracting crucial patterns and potential incongruities. 3) Engaging the agents in a structured debate and arbitration process to formulate a robust, well-reasoned final judgment on the article's authenticity.

Our primary contributions are summarized as follows:
\begin{itemize}
    \item We introduce a pioneering multi-agent framework for zero-shot multimodal fake news detection that uniquely harnesses retrieval-augmented reasoning, thereby removing the dependency on labeled data for newsworthy events.
    \item We propose a robust Bifurcated Evidential Reasoning mechanism, where agents perform dual-directional analysis on retrieved articles to build a comprehensive understanding of disinformation tactics and narrative structures.
    \item We design a Multi-Agent Collaborative Debate process, with specialized analyst and arbiter roles, which elevates the reliability and interpretability of decision-making by resolving analytical conflicts through structured discourse.
    \item Comprehensive experiments on benchmark fake news datasets confirm that our proposed MRAFnd framework substantially outperforms state-of-the-art zero-shot baselines.
\end{itemize}

\section{Related Work}

\subsection{Multimodal Fake News Detection}
With the development of multimodal technology \cite{lu2025dmmd4sr,cui2025multi,cui2025diffusion}, multimodal fake news detection has gradually become the mainstream. Multimodal fake news detection aims to identify inconsistencies across modalities \cite{liu2023tmac}. Methodologies have evolved from simple feature fusion of VGG and BERT embeddings \cite{khattar2019mvae} to more sophisticated fusion techniques and graph-based networks for event-level analysis \cite{lu2025dammfnd, ma2024event}. Recently, Large Language Models (LLMs) have been employed in specialized frameworks like TELLER and SNIFFER to enhance reasoning capabilities \cite{liu2024teller, qi2024sniffer}. While these methods show promise, they often require fine-tuning or labeled examples, which are unavailable for emerging disinformation campaigns. MRAFnd differs by operating in a truly zero-shot setting, leveraging retrieval from an unannotated corpus to analyze disinformation patterns without any model training.

\subsection{LLM-Based Multi-Agent Frameworks}
LLMs have evolved into autonomous agents capable of complex reasoning and planning \cite{wang2023voyager}, leading to the development of multi-agent systems where agents collaborate to solve complex problems \cite{park2023generative}. However, many such systems depend on environmental feedback, which is incompatible with zero-shot classification as it would reveal the ground-truth label. MRAFnd's architecture is fundamentally different: it is feedback-free, orchestrating agent collaboration through structured analysis and debate to enable robust zero-shot reasoning without any supervisory signals.

\begin{figure*}[t!]
    \centering
    \includegraphics[width=\linewidth]{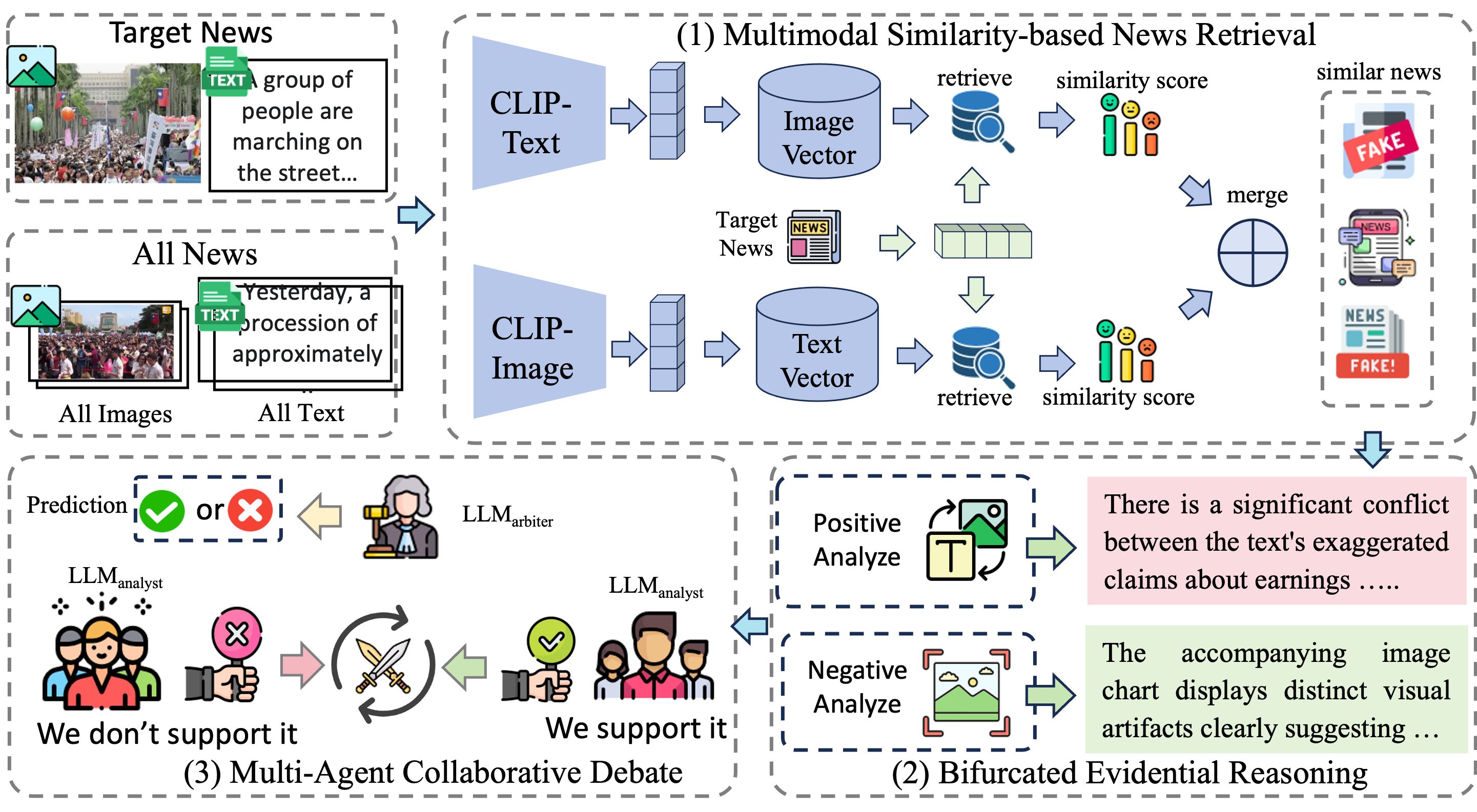} 
    \caption{An overview of our proposed framework, MRAFnd, for zero-shot multimodal fake news detection. The process is composed of three primary stages: (1) Multimodal Similarity-based News Retrieval, (2) Bifurcated Evidential Reasoning, and (3) Multi-Agent Collaborative Debate. Note: Module names in the figure should be mentally mapped to the new names used in this paper.}
    \label{fig:framework}
\end{figure*}

\section{Methodology}

\subsection{Overview}

\paragraph{Problem Formulation}
We define a multimodal news collection as a set of articles, where each article $N$ is represented as a tuple $\{\mathcal{V}, \mathcal{T}\}$, containing a visual element $\mathcal{V}$ (image) and a textual element $\mathcal{T}$ (text). Our task is zero-shot fake news detection, where the objective is to classify articles in a test set $S_{\text{test}}$ as either authentic or fabricated. This is achieved by utilizing an unlabeled reference set of news articles, $S_{\text{ref}}$, as a source of contextual evidence. Our method is entirely gradient-free, obviating the need for model training or fine-tuning.

To address the data scarcity inherent in detecting novel fake news, we introduce \textbf{MRAFnd}, a multi-agent framework conceived to simulate a collaborative verification workflow. MRAFnd orchestrates multiple LLM agents across three sequential stages: 1) \textbf{Multimodal Similarity-based News Retrieval} (\S\ref{sec:retrieval}), which finds and collects contextually similar news articles from the unlabeled reference set $S_{\text{ref}}$. 2) \textbf{Bifurcated Evidential Reasoning} (\S\ref{sec:analysis}), where agents perform a dual-directional analysis to extract key analytical patterns from the retrieved evidence. 3) \textbf{Multi-Agent Collaborative Debate} (\S\ref{sec:deliberation}), in which specialized analyst and arbiter agents discuss the findings to reach a final, well-founded verdict. A schematic of MRAFnd is depicted in Figure~\ref{fig:framework}.

\subsection{Multimodal Similarity-based News Retrieval}
\label{sec:retrieval}

Disinformation campaigns, despite their continuous evolution, often display recurrent thematic and structural motifs \cite{zhou2020survey}. Inspired by retrieval-augmented generation \cite{lewis2020retrieval}, our initial stage is dedicated to retrieving news articles from $S_{\text{ref}}$ that are analogous to the target article. These retrieved items function as critical evidence, supplying the context required to evaluate the target's veracity.

For a given target news article \(N_{target} = \{\mathcal{V}_{target}, \mathcal{T}_{target}\}\), we first generate its multimodal embedding by integrating its visual and textual features:
\begin{equation}
\label{eq:encode}
\mathbf{E} = \lambda_v \cdot \mathbf{V_{enc}}(\mathcal{V}) + \lambda_t \cdot \mathbf{T_{enc}}(\mathcal{T}),
\end{equation}
where $\mathbf{E}$ denotes the final multimodal embedding. $\mathbf{V_{enc}}(\cdot)$ and $\mathbf{T_{enc}}(\cdot)$ are pre-trained visual and textual encoders, respectively. The coefficients $\lambda_v$ and $\lambda_t$ are fixed hyperparameters that modulate the influence of each modality. We apply this encoding procedure to the target article and all articles in the reference set $S_{\text{ref}}$.

To pinpoint the most pertinent evidence, we calculate the cosine similarity between the embedding of the target article and those of all articles in the reference set:
\begin{equation}
s = \text{sim}(\mathbf{E}_{\text{target}}, \mathbf{E}_{\text{ref}}) = \frac{\mathbf{E}_{\text{target}} \cdot \mathbf{E}_{\text{ref}}}{\|\mathbf{E}_{\text{target}}\| \|\mathbf{E}_{\text{ref}}\|},
\end{equation}
where $\mathbf{E}_{\text{target}}$ and $\mathbf{E}_{\text{ref}}$ are the embeddings for the target and a reference article, respectively. Based on these similarity scores, we select the top-$K$ articles from $S_{\text{ref}}$ to constitute the evidence set $N_{\text{evidence}}$:
\begin{equation}
N_{\text{evidence}} = \{ N_{\text{ref}} \mid s \in \text{Top}_K(\{ \text{sim}(\mathbf{E}_{\text{target}}, \mathbf{E}_{ref}) \mid N_{ref} \in S_{ref} \}) \}.
\end{equation}
This evidence set, comprising the $K$ most similar yet unlabeled news articles, is subsequently forwarded to the next stage for detailed analysis.

\subsection{Bifurcated Evidential Reasoning}
\label{sec:analysis}
The retrieved evidence set, $N_{\text{evidence}}$, offers valuable context. However, since the articles are unlabeled, presenting them directly to a model could introduce noise and ambiguity. To mitigate this, we introduce a structured reasoning stage where an LLM agent systematically extracts relevant patterns and potential indicators of misinformation. This process is executed from two opposing directions to ensure a comprehensive and impartial analysis.

\subsubsection{Affirmative Pattern Review}
The reasoning agent processes the $K$ evidence articles in their original sequence. At each step $i$, the agent evaluates the $i$-th article in light of the analytical insights gathered from the preceding $i-1$ steps. This iterative approach enables the agent to construct a cumulative understanding of narrative patterns, visual motifs, and potential inconsistencies, thereby building an affirmative case based on the evidence flow. This process is formalized as:
\begin{equation}
\mathcal{P}_{\text{affirm}, i} = \text{LLM}_\text{reasoner}(N_{\text{evidence}, i}, \mathcal{P}_{\text{affirm}, i-1}, \Pi_{\text{reasoner}}),
\end{equation}
where \(\mathcal{P}_{\text{affirm}, i}\) is the set of distilled affirmative patterns after processing the $i$-th article. $N_{\text{evidence}, i}$ is the $i$-th article from the evidence set, and $\text{LLM}_\text{reasoner}$ is an LLM agent guided by a Chain-of-Thought prompt $\Pi_{\text{reasoner}}$ designed to elicit detailed reasoning about indicators of fake news.

\subsubsection{Negational Pattern Review}
A potential drawback of a purely sequential analysis is order dependency, where articles encountered earlier may unduly influence the final summary. To counteract this bias, we implement a complementary negational review. Upon completing the affirmative pass, the reasoning agent re-processes the same evidence set but in reverse order. This pass is designed to challenge or negate the initial findings by approaching the evidence from an opposing viewpoint.
\begin{equation}
\mathcal{P}_{\text{negate}, i} = \text{LLM}_\text{reasoner}(N_{\text{evidence}, K+1-i}, \mathcal{P}_{\text{negate}, i-1}, \Pi_{\text{reasoner}}),
\end{equation}
where \(\mathcal{P}_{\text{negate}, i}\) represents the accumulated findings from the negational review. This dual-perspective approach ensures that each piece of evidence is assessed from two distinct contextual standpoints, yielding two comprehensive sets of findings: \(\mathcal{P}_{\text{affirm}, K}\) and \(\mathcal{P}_{\text{negate}, K}\).

\subsection{Multi-Agent Collaborative Debate}
\label{sec:deliberation}
In the final stage, we institute a multi-agent debate to synthesize the analytical findings and render a definitive judgment. This stage engages two types of agents: Analyst Agents and an Arbiter Agent.

First, two Analyst Agents are assigned the task of formulating independent assessments. Each agent is provided with the target news article and one of the two sets of findings (affirmative or negational) from the preceding stage. They each generate a detailed report outlining their reasoning and a final verdict (real or fake):
\begin{equation}
\mathcal{R}_{\text{affirm}} = \text{LLM}_{\text{analyst}}(\mathcal{P}_{\text{affirm}, K}, \mathcal{V}_{\text{target}}, \mathcal{T}_{\text{target}}),
\end{equation}
\begin{equation}
\mathcal{R}_{\text{negate}} = \text{LLM}_{\text{analyst}}(\mathcal{P}_{\text{negate}, K}, \mathcal{V}_{\text{target}}, \mathcal{T}_{\text{target}}),
\end{equation}
where \(\mathcal{R}_{\text{affirm}}\) and \(\mathcal{R}_{\text{negate}}\) are the reasoned reports from the two analyst agents, respectively. This step ensures the development of two independent lines of reasoning from the bifurcated analysis.

Next, the verdicts from the two reports are compared. If the analysts concur, their shared decision is accepted as the final outcome. In instances of disagreement, a third Arbiter Agent is summoned to resolve the conflict. The arbiter examines the reasoning of both analysts to make a final, decisive judgment:
\begin{equation}
\mathcal{R}_\text{final} = 
\begin{cases} 
\mathcal{R}_{\text{affirm}} & \text{if } \text{verdict}(\mathcal{R}_{\text{affirm}}) = \text{verdict}(\mathcal{R}_{\text{negate}}) \\
\text{LLM}_\text{arbiter}(\mathcal{R}_{\text{affirm}}, \mathcal{R}_{\text{negate}}, \mathcal{V}_{\text{target}}, \mathcal{T}_{\text{target}}) & \text{otherwise}
\end{cases}, 
\end{equation}
where \(\text{LLM}_\text{arbiter}\) meticulously weighs the arguments from both perspectives before making the final determination. This debate mechanism enhances the robustness and reliability of the final decision by incorporating multiple viewpoints and establishing a structured protocol for conflict resolution.

\section{Experiments}
\label{sec:experiments}

In this section, we present a comprehensive suite of experiments to assess the efficacy of our proposed MRAFnd framework. Our evaluation is structured to address the following research questions:
\begin{itemize}
    \item \textbf{RQ1:} How does MRAFnd's performance compare to state-of-the-art multimodal and LLM-based fake news detection methods in a zero-shot context?
    \item \textbf{RQ2:} What is the individual contribution of each core component within the MRAFnd framework, namely news retrieval, bifurcated reasoning, and collaborative debate?
    \item \textbf{RQ3:} How sensitive is MRAFnd's performance to its primary hyperparameters, specifically the number of retrieved articles and the rounds of debate?
    \item \textbf{RQ4:} What is the inference efficiency of MRAFnd concerning token consumption and processing time relative to other LLM-based methods?
    \item \textbf{RQ5:} How robust is MRAFnd to noise and imperfections during the evidence retrieval phase?
\end{itemize}

\subsection{Experimental Settings}

\subsubsection{Datasets}
We evaluate our framework on three widely-adopted public benchmark datasets for multimodal fake news detection:
\begin{itemize}
    \item \textbf{Weibo} \cite{wang2018eann}: A classic Chinese dataset sourced from the Weibo social media platform, serving as a standard benchmark for this task.
    \item \textbf{Weibo-21} \cite{nan2021mdfend}: A more contemporary and challenging Chinese dataset, also from Weibo, featuring multi-domain news content that tests a model's generalization capabilities.
    \item \textbf{GossipCop} \cite{shu2020fakenewsnet}: A real-world English dataset centered on celebrity and entertainment news, curated by professional journalists and known for its subtle instances of misinformation.
\end{itemize}

\subsubsection{Baselines}
We compare MRAFnd against three classes of state-of-the-art baselines in a zero-shot setting:
\begin{itemize}
    \item \textbf{General-purpose LLMs}, which process only the textual component of the news: GPT-4o and Gemini-1.5-Flash.
    \item \textbf{General-purpose MLLMs}, which can interpret both text and images: LLaVA-1.5-7B \cite{liu2024improved}, InstructBLIP-7B \cite{instructblip}, and MiniGPT-v2-7B \cite{chen2023minigpt}.
    \item \textbf{LLM-based FND Methods}, which employ advanced reasoning for fact-checking: TELLER \cite{liu2024teller}, SNIFFER \cite{qi2024sniffer}, and FactAgent \cite{li2024large}.
\end{itemize}

\subsubsection{Implementation Details}
All experiments were performed on a server with NVIDIA A100 GPUs. For our MRAFnd framework, we use LLaVA-1.5-7B as the backbone for all agents, chosen for its strong multimodal reasoning and efficient architecture. The visual encoder and text encoder are the pre-trained ViT-L/14 from CLIP \cite{radford2021learning}. For evidence retrieval, we set the number of retrieved articles $K=5$ and the number of debate rounds $R=2$, based on the analysis in Section \ref{sec:param_analysis}. We evaluate all methods using two standard metrics: \textbf{Accuracy} and \textbf{Macro-F1} score.

\subsection{Main Results (RQ1)}
\label{sec:main_results}

Table \ref{tab:main_results} displays the main experimental outcomes on the three benchmark datasets. Our proposed framework, MRAFnd, consistently achieves the highest performance across all datasets and metrics, affirming its superior efficacy in the zero-shot multimodal fake news detection task.

\begin{table*}[t!]
\centering
\caption{Main results of zero-shot multimodal fake news detection. We report Accuracy (\%) and Macro-F1 (\%). The best results are in \textbf{bold}, and the second-best are \underline{underlined}. All improvements are statistically significant ($p < 0.05$).}
\label{tab:main_results}
\resizebox{\linewidth}{!}{%
\begin{tabular}{@{}l l cc cc cc@{}}
\toprule
\multirow{2}{*}{\textbf{Category}} & \multirow{2}{*}{\textbf{Method}} & \multicolumn{2}{c}{\textbf{Weibo}} & \multicolumn{2}{c}{\textbf{Weibo-21}} & \multicolumn{2}{c}{\textbf{GossipCop}} \\ \cmidrule(lr){3-4} \cmidrule(lr){5-6} \cmidrule(lr){7-8}
& & Accuracy & Macro-F1 & Accuracy & Macro-F1 & Accuracy & Macro-F1 \\ \midrule \midrule
\multirow{2}{*}{General-purpose LLMs} 
& GPT-4o & 75.31 & 75.15 & 78.40 & 78.31 & 67.28 & 65.04 \\
& Gemini-1.5-Flash & 74.84 & 74.68 & 77.65 & 77.52 & 66.53 & 64.18 \\ \midrule
\multirow{3}{*}{General-purpose MLLMs} 
& LLaVA-1.5-7B \cite{liu2024improved} & 78.16 & 78.02 & 80.27 & 80.19 & 68.31 & 66.85 \\
& InstructBLIP-7B \cite{instructblip} & 77.21 & 77.05 & 79.52 & 79.46 & 67.73 & 66.12 \\
& MiniGPT-v2-7B \cite{chen2023minigpt} & 77.86 & 77.74 & 79.84 & 79.80 & 68.07 & 66.53 \\ \midrule
\multirow{3}{*}{LLM-based FND Methods} 
& TELLER \cite{liu2024teller} & 81.25 & 81.13 & 83.10 & 83.02 & 70.89 & 69.84 \\
& SNIFFER \cite{qi2024sniffer} & 80.83 & 80.75 & 82.57 & 82.49 & 70.18 & 69.05 \\
& FactAgent \cite{li2_2024_large} & \underline{82.14} & \underline{82.06} & \underline{83.96} & \underline{83.85} & \underline{71.39} & \underline{70.21} \\ \midrule
\textbf{Ours} & \textbf{MRAFnd (LLaVA-1.5-7B)} & \textbf{84.75} & \textbf{84.69} & \textbf{86.31} & \textbf{86.25} & \textbf{73.55} & \textbf{72.48} \\ \bottomrule
\end{tabular}%
}
\end{table*}

\textbf{MRAFnd substantially outperforms all baselines by a significant margin.} On the challenging Weibo-21 dataset, for example, MRAFnd achieves an accuracy of 86.31\%, surpassing the strongest baseline, FactAgent, by 2.35\%. This result underscores the efficacy of our core design: by retrieving and analyzing a collection of evidential articles, MRAFnd transcends isolated analysis. It successfully identifies the recycled narrative structures and manipulation tactics prevalent in disinformation campaigns—a crucial capability that other methods lack.

\textbf{LLM-based FND methods exhibit stronger performance than general-purpose models.} Specialized methods like FactAgent and TELLER, which incorporate fact-checking pipelines, outperform general-purpose MLLMs. This is anticipated, as their structured reasoning workflows are better suited to the fake news detection task. However, their dependence on analyzing the target article in isolation constrains their ability to contextualize it within broader disinformation trends.

\textbf{Multimodal models show a clear advantage over text-only LLMs.} MLLMs such as LLaVA consistently outperform text-only models like GPT-4o. This confirms the value of incorporating visual information, as inconsistencies between image and text often serve as key indicators of fake news. Nevertheless, without a mechanism to harness external context, their performance remains constrained.

\subsection{Ablation Study (RQ2)}
To assess the contribution of each component in MRAFnd, we conduct an ablation study with the following variants:
\begin{itemize}
    \item \textbf{w/o Retrieval}: Removes the evidence retrieval stage, analyzing the target news in isolation, similar to standard MLLM approaches.
    \item \textbf{w/o Bifurcated Reasoning}: Replaces the dual-directional analysis with a single-pass review of the retrieved evidence.
    \item \textbf{w/o Affirmative Distillation}: Uses only the results from the negational pattern review.
    \item \textbf{w/o Negational Review}: Uses only the results from the Affirmative Pattern Review.
    \item \textbf{w/o Collaborative Debate}: Eliminates the multi-agent debate stage. In cases of disagreement between the two initial analyses, it defaults to the conclusion from the affirmative pass.
\end{itemize}

\begin{figure*}[t!]
    \centering
    \includegraphics[width=\linewidth]{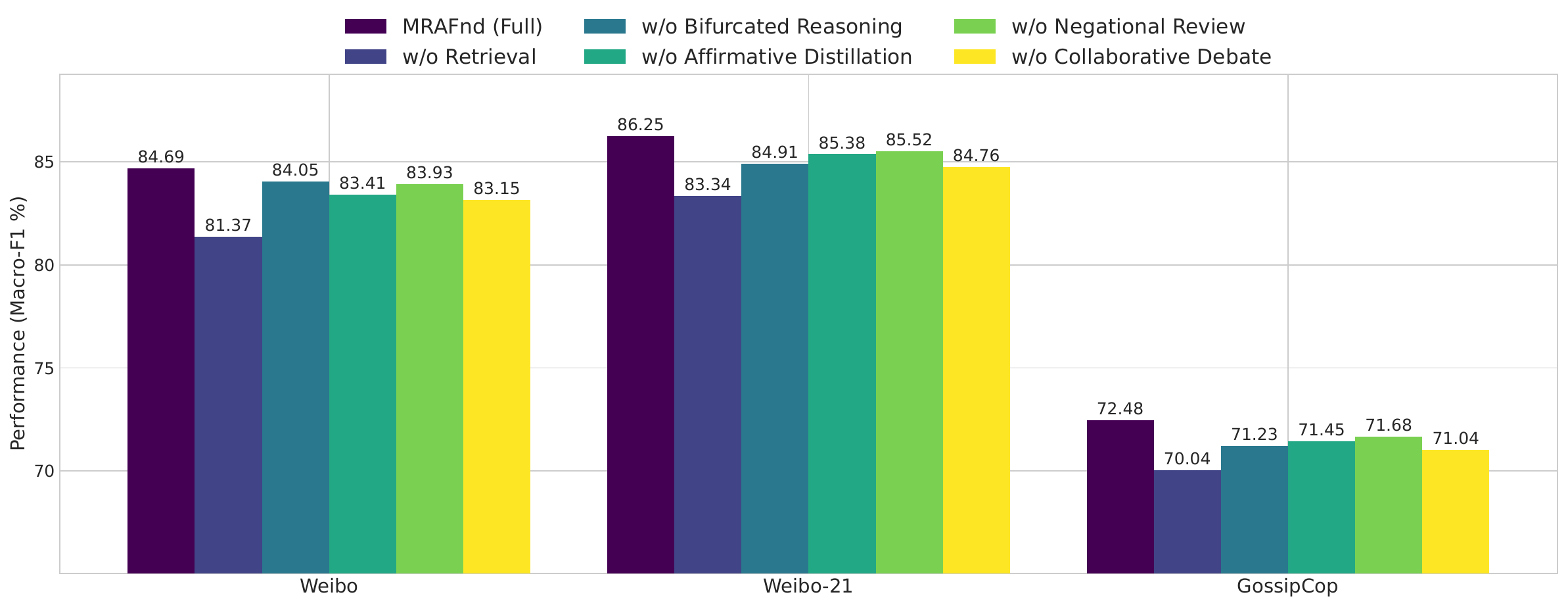} 
    \caption{Ablation study of MRAFnd on three datasets. Performance is reported as Macro-F1 score.}
    \label{fig:ablation}
\end{figure*}

Figure \ref{fig:ablation} presents the results of the ablation study. The removal of any component results in a performance decline, confirming the integral role of each part of our framework. The most substantial performance drop is observed in the \textbf{w/o Retrieval} configuration, where the Macro-F1 score decreases by 3.32 percentage points on Weibo. This clearly illustrates that retrieval-augmented context is the cornerstone of MRAFnd's success. Removing the \textbf{Bifurcated Reasoning} also leads to a notable decrease, emphasizing the importance of mitigating order bias and achieving a holistic understanding of the evidence. Importantly, even when key components are ablated (e.g., w/o Collaborative Debate), our framework variants still generally outperform the strongest baseline, underscoring the resilience of the core retrieval-augmented reasoning paradigm.

\subsection{Parameter Sensitivity Analysis (RQ3)}
\label{sec:param_analysis}
We examine the impact of two critical hyperparameters: the number of retrieved articles ($K$) and the number of debate rounds ($R$).

\begin{figure*}[t!]
    \centering
    \includegraphics[width=0.9\linewidth]{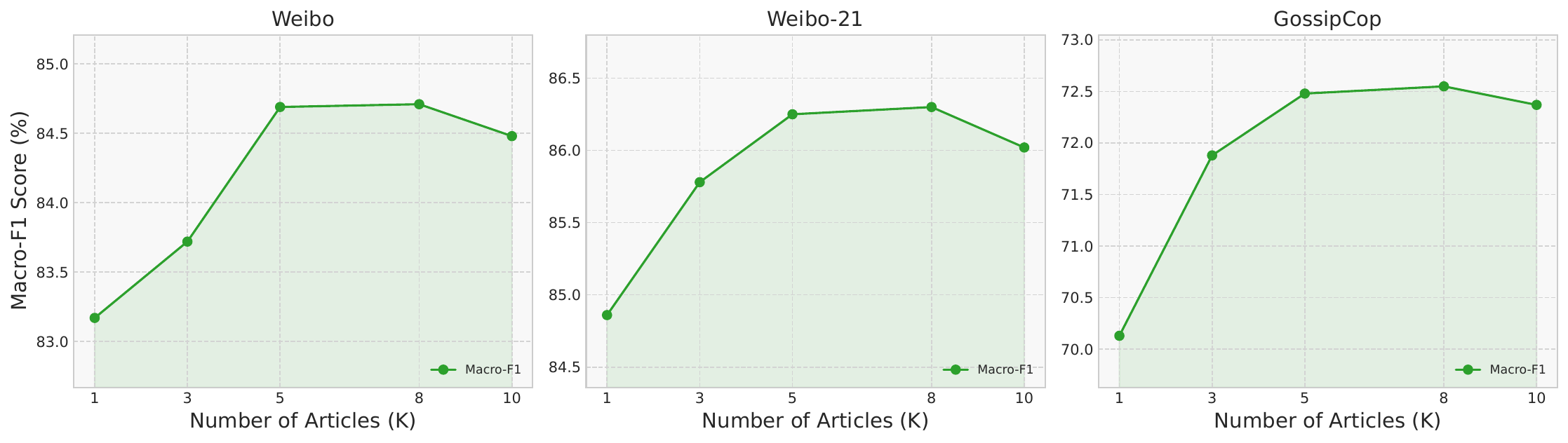} 
    \caption{Performance of MRAFnd with a varying number of retrieved articles ($K$).}
    \label{fig:param_k}
\end{figure*}

\begin{figure*}[t!]
    \centering
    \includegraphics[width=0.9\linewidth]{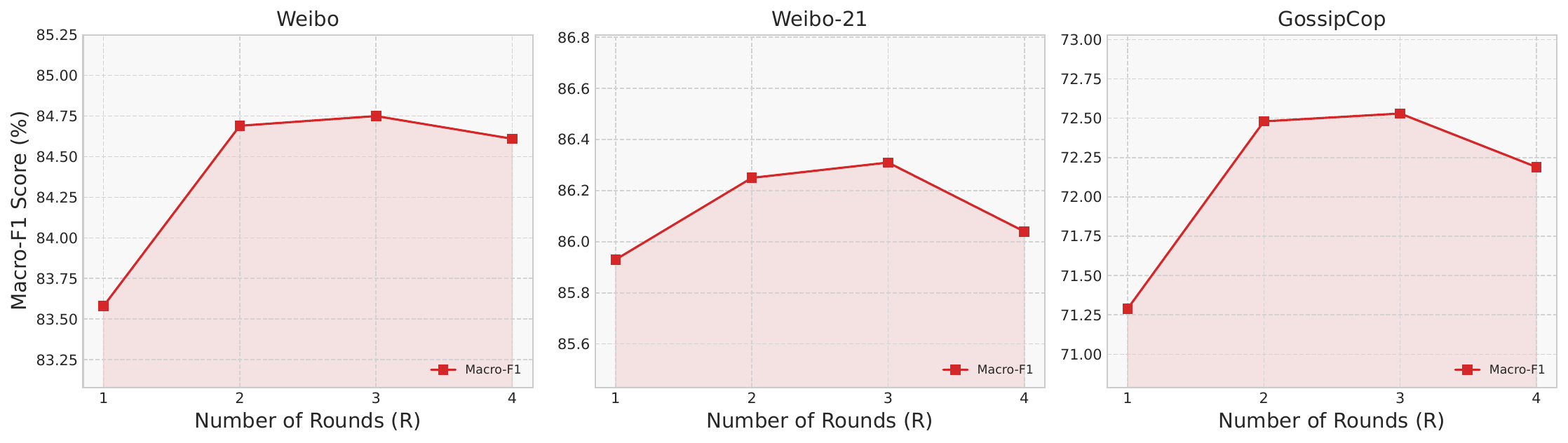} 
    \caption{Performance of MRAFnd with a varying number of deliberation rounds ($R$).}
    \label{fig:param_r}
\end{figure*}

\textbf{Effect of Retrieved Articles ($K$):} As depicted in Figure \ref{fig:param_k}, performance generally enhances as $K$ is increased from 1 to 5. This is because a larger evidence pool provides a richer context for identifying malicious patterns. However, for $K$ greater than 5, performance starts to plateau or decline slightly. This suggests that an excessive number of retrieved articles may introduce noise from less relevant instances, potentially hindering the reasoning process. We therefore select $K=5$ as the optimal value.

\textbf{Effect of Debate Rounds ($R$):} Figure \ref{fig:param_r} illustrates the effect of the number of debate rounds between the Analyst and Arbiter agents. Transitioning from $R=1$ (a single decision) to $R=2$ (one round of debate and refinement) produces a tangible performance improvement across all datasets. This shows that enabling agents to challenge and refine initial conclusions helps resolve ambiguities and leads to a more robust final verdict. Performance stabilizes at $R=3$, indicating that two rounds are sufficient to converge on a well-reasoned decision. We choose $R=2$ for an optimal balance between performance and efficiency.

\subsection{Inference Efficiency Analysis (RQ4)}
We compare the inference efficiency of MRAFnd with other LLM-based FND methods, measuring the average number of tokens consumed and the average wall-clock time required per news article on the Weibo dataset.

\begin{table}[H]
\centering
\caption{Inference efficiency comparison on the Weibo dataset.}
\label{tab:efficiency}
\begin{tabular}{@{}lcc@{}}
\toprule
\textbf{Method} & \textbf{Avg. Tokens / Sample} & \textbf{Avg. Time / Sample (s)} \\ \midrule
TELLER          & $\sim$1.8k                    & 10.2                          \\
SNIFFER         & $\sim$2.1k                    & 12.5                          \\
FactAgent       & $\sim$4.5k                    & 23.8                          \\ \midrule
\textbf{MRAFnd (Ours)}  & $\sim$3.2k                    & 14.7                          \\ \bottomrule
\end{tabular}
\end{table}

As shown in Table \ref{tab:efficiency}, MRAFnd uses fewer tokens than the complex, multi-step FactAgent framework, demonstrating superior token efficiency. While its token consumption is higher than simpler methods like TELLER, this is an expected trade-off for the comprehensive analysis of retrieved evidence. Crucially, MRAFnd's inference time is only slightly longer than SNIFFER's and markedly faster than FactAgent's. This efficiency is achieved by parallelizing the analysis of different evidence articles and employing a lightweight multi-agent debate. This analysis confirms that MRAFnd achieves its state-of-the-art performance with a practical and manageable computational footprint.

\subsection{Robustness to Retrieval Noise (RQ5)}
To evaluate the framework's resilience, we simulate a noisy retrieval environment. We corrupt the retrieved evidence set by substituting a percentage (from 10\% to 50\%) of the top-$K$ articles with randomly selected articles from the reference corpus.

\begin{figure*}[t!]
    \centering
    \includegraphics[width=0.8\linewidth]{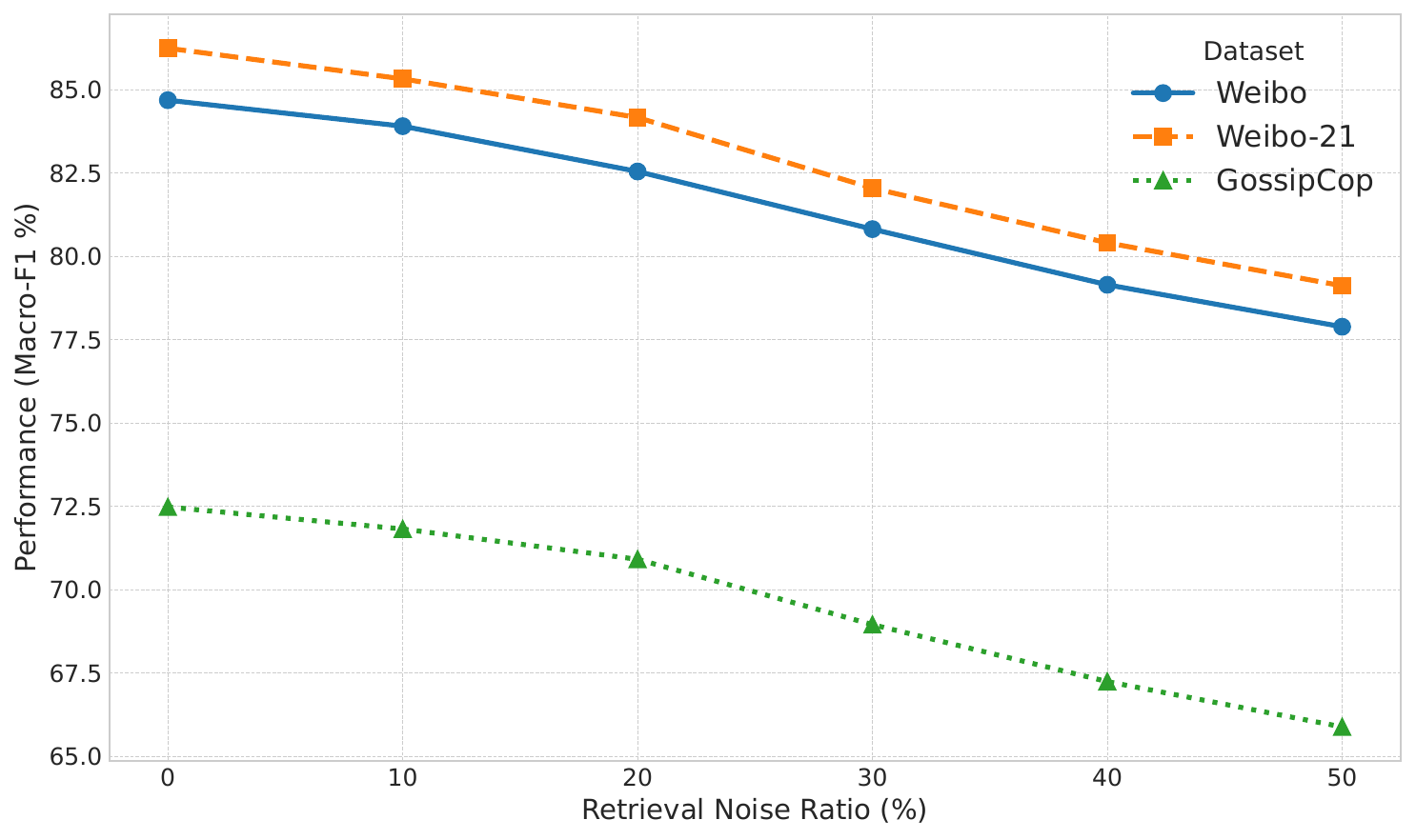} 
    \caption{Performance of MRAFnd under varying levels of retrieval noise. The noise ratio indicates the percentage of retrieved articles replaced with random ones.}
    \label{fig:robustness}
\end{figure*}

Figure \ref{fig:robustness} illustrates that MRAFnd's performance degrades gracefully as the noise level increases. Even when 30\% of the retrieved evidence is irrelevant, MRAFnd's Macro-F1 score drops by only about 4-5\% on average. Its performance remains competitive even with 50\% noise. This highlights the robustness of our Bifurcated Evidential Reasoning and Multi-Agent Collaborative Debate stages, which can effectively distill meaningful signals and filter out distractions from a partially corrupted evidence set—a vital capability for real-world deployments where retrieval systems are inherently imperfect.

\subsection{Robustness to MLLM Backbone}
\begin{figure*}[h!]
    \centering
    \includegraphics[width=0.9\textwidth]{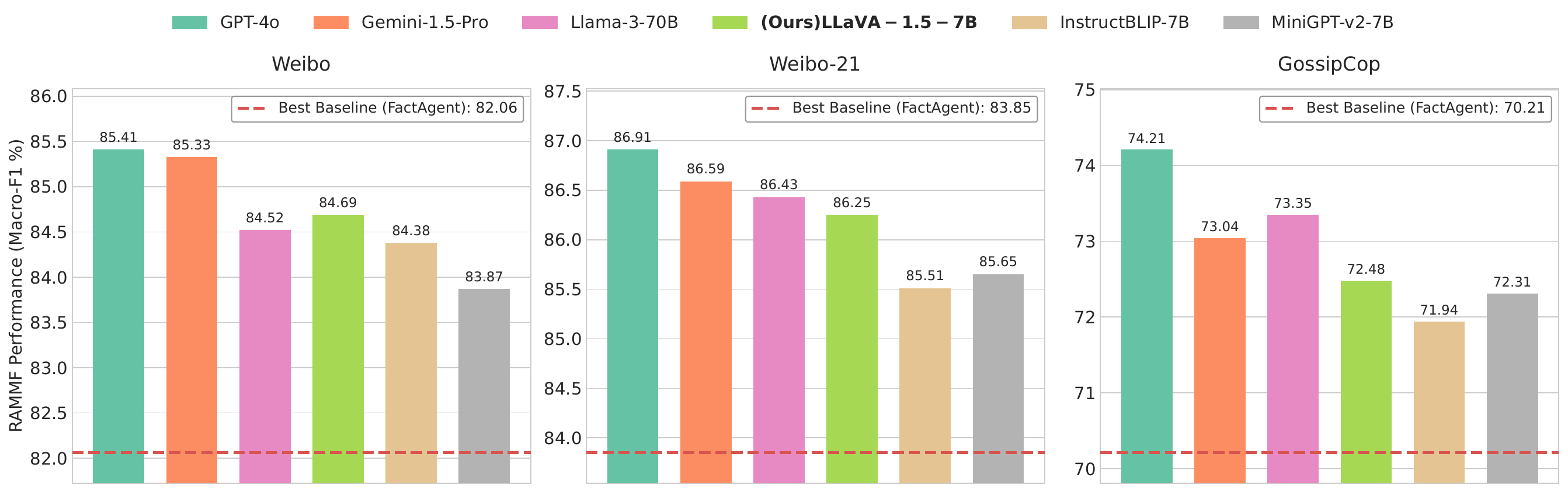}
    \caption{Performance (Macro-F1 \%) of the MRAFnd framework with different MLLM backbones on the three datasets.}
    \label{fig:mllm_robustness}
\end{figure*}

To confirm that MRAFnd's effectiveness stems from its architectural design rather than the capability of a single model, we assess its performance with various MLLM backbones. We substituted the agents' core model with several leading MLLMs, including large proprietary models (GPT-4o, Gemini-1.5-Pro) and powerful open-source alternatives (Llama-3-70B, InstructBLIP-7B). The Macro-F1 scores are presented in Figure~\ref{fig:mllm_robustness}.

The results unequivocally demonstrate MRAFnd's robustness. The framework consistently outperforms the strongest baseline (FactAgent) across all tested MLLMs. As anticipated, larger models like GPT-4o deliver the highest performance. However, the performance reduction when using smaller models is notably gradual. Remarkably, MRAFnd powered by the lightweight LLaVA-1.5-7B not only significantly surpasses FactAgent but also remains competitive with much larger models. This provides strong evidence that MRAFnd's agentic workflow—retrieving evidence, analyzing it from dual perspectives, and debating the outcomes—is the primary driver of its success. This insight is crucial, as it confirms that \textbf{MRAFnd can be deployed effectively using smaller, open-source models, offering a state-of-the-art solution that is both powerful and practical for real-world applications}.

\section{Conclusion}
In this paper, we proposed \textbf{MRAFnd}, a novel retrieval-augmented multi-agent framework for zero-shot multimodal fake news detection. Our framework mimics a collaborative analytical process by integrating \textbf{Multimodal News Retrieval} to gather evidence, \textbf{Bifurcated Evidential Reasoning} to mitigate analytical bias, and a \textbf{Multi-Agent Collaborative Debate} to synthesize findings into a robust verdict. Extensive experiments confirm MRAFnd significantly outperforms state-of-the-art baselines across three benchmark datasets. Future work will address the framework's dependency on retrieval quality by enhancing the retrieval mechanism's resilience to noise and modal inconsistencies, thereby further improving detection reliability.

\bibliography{ref}
\bibliographystyle{splncs04}

\end{document}